\documentclass[letterpaper, 10 pt, conference]{ieeeconf}  

\IEEEoverridecommandlockouts                              

\usepackage{graphics} 
\usepackage{epsfig} 
\usepackage{amsmath} 
\usepackage{amssymb}  
\usepackage{cite}
\usepackage{graphicx}
\usepackage{subcaption}
\usepackage{comment}

\title{\LARGE \bf
Grounding Commands for Autonomous Vehicles via Layer Fusion \\with Region-specific Dynamic Layer Attention
}

\author{Hou Pong Chan$^{*1}$, Mingxi Guo$^{*1}$, and Cheng-Zhong Xu$^{\dagger 1}$
\thanks{$^*$Both authors contributed equally to this research.}
\thanks{$\dagger$Corresponding author. }
\thanks{$^{1}$Department of Computer and Information Science, University of Macau, Macau SAR, China.
        Email: {\tt\small \{hpchan, mc05415, czxu\}@um.edu.mo}}%
}

\begin{document}

\maketitle
\thispagestyle{empty}
\pagestyle{empty}

\begin{abstract}
Grounding a command to the visual environment is an essential ingredient for interactions between autonomous vehicles and humans. In this work, we study the problem of language grounding for autonomous vehicles, which aims to localize a region in a visual scene according to a natural language command from a passenger. Prior work only employs the top layer representations of a vision-and-language pre-trained model to predict the region referred to by the command. However, such a method omits the useful features encoded in other layers, and thus results in inadequate understanding of the input scene and command. To tackle this limitation, we present the first layer fusion approach for this task. Since different visual regions may require distinct types of features to disambiguate them from each other, we further propose the region-specific dynamic (RSD) layer attention to adaptively fuse the multimodal information across layers for each region. Extensive experiments on the Talk2Car benchmark demonstrate that our approach helps predict more accurate regions and outperforms state-of-the-art methods. 
\end{abstract}

\section{Introduction}
\begin{figure}[t]
\centering
\begin{tabular}{p{0.96\columnwidth}}
\includegraphics[width=\linewidth]{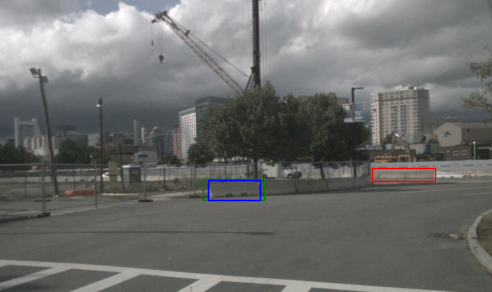}\\
\textbf{Command: }find a parking spot near the \textit{first concrete barrier}. 
\end{tabular}
\caption{
A sample image and the associated natural language command. The blue box indicates the ground-truth region referred to by the command. The red box shows the prediction by the UNITER model~\cite{DBLP:conf/eccv/UNITER20} when using the top layer representations to the compute matching scores of regions. 
The green box is predicted by the UNITER model after enhanced by our RSD layer attention approach. 
}
\label{fig:intro-case}
\end{figure}

In recent years, autonomous driving systems achieve significant progress due to the advances in sensing technologies and deep neural networks. 
However, self-driving requires a very high level of trust from users and it will only be adopted if it creates a better human experience~\cite{MITFriman2012lecture}. 
Prior literature~\cite{DBLP:conf/cvpr/KimMRDC20} reveals that enabling users to issue commands to autonomous vehicles via natural language can improve user experience and acceptance. 
In order to perform the action specified by a command (e.g., follow a particular car or stop at a specific parking spot), a vehicle needs to understand the semantic correspondence between the natural language command and the visual environment. This problem is formalized as the task of \textit{language grounding for autonomous vehicles}~\cite{DBLP:conf/emnlp/talk2car19}: given an image and a natural language command to a vehicle, the goal is to locate the region in the image that is referred to by the command. We give a sample of image, command, and the referred region in Figure~\ref{fig:intro-case}. 


Existing work typically uses an object detector to extract region proposals from the input image and casts the task as selecting the best-matched region based on the command. 
State-of-the-art (SOTA) methods~\cite{DBLP:conf/eccv/DaiLDS20} apply a vision-and-language (V\&L) pre-trained model to summarize the region proposals and command words into contextualized representations through multiple Transformer~\cite{DBLP:conf/nips/Transformer17} encoder layers, and then use the top layer representations to predict the matching scores of regions. 
Several studies~\cite{DBLP:conf/naacl/Elmo18,DBLP:conf/acl/BertPipeline19} show that the representations learned by different encoder layers embed different types of surface and semantic features. 
However, SOTA language grounding methods only utilize the top layer representations to compute the matching scores. 
In consequence, these methods ignore the useful features inside the representations in other layers, which may lead to insufficient comprehension of the input scene and command. 
For instance, Figure~\ref{fig:intro-case} shows a V\&L pre-trained model, UNITER~\cite{DBLP:conf/eccv/UNITER20}, outputs an incorrect region when using the top layer representations to predict the matching scores of regions. 




To effectively utilize the features embedded in different encoder layers, we propose \textit{the first encoder layer fusion approach for the language grounding task}. For each region proposal, our approach fuses its representations across all encoder layers in a V\&L pre-trained model. After that, we feed the fused representation to an output layer to predict its matching score. Our work investigates the layer attention technique~\cite{DBLP:conf/naacl/Elmo18,DBLP:conf/emnlp/BapnaCFCW18} for encoder layer fusion since it has good interpretability and we can directly assess the contribution made by each encoder layer. 

Current layer attention methods assign a static set of normalized attention weights to encoder layers regardless of the input. 
Then they use the attention weights to aggregate the encoder representations over layers by weighted sum. 
However, in the language grounding problem, different visual regions may require distinct types of linguistic and visual features to determine whether they match the command. As an example, for the trees in Figure~\ref{fig:intro-case}, a model only needs the object category information to reject them as the correction region. Whereas for the concrete barriers in the figure, a model requires both the object category and position features to determine whether they match the command. 

To address the above drawback of existing layer attention methods, we further propose the \textbf{region-specific dynamic (RSD) layer attention} mechanism, which dynamically computes a new set of layer attention weights for each individual image region. The attention weights are learned by a neural module based on the representations of the regions. 
Thus, our method can adaptively determine the importance of different encoder layers and acquire appropriate features for each visual region. 
Figure~\ref{fig:intro-case} illustrates that our RSD layer attention helps predict a more accurate region\footnote{We also provide a demo of our approach in the attached video.}. 

Comprehensive empirical studies are conducted on the Talk2Car~\cite{DBLP:conf/emnlp/talk2car19} benchmark. 
We apply our layer fusion approach to enhance two recent V\&L pre-trained models, UNITER~\cite{DBLP:conf/eccv/UNITER20} and LXMERT~\cite{DBLP:conf/emnlp/TanB19}. 
Experiment results demonstrate that our approach consistently improves the accuracy of both the UNITER and LXMERT models and outperforms the SOTA methods in this task. 
Moreover, our proposed RSD layer attention also achieves better performance than existing layer fusion methods. We then examine how our method distributes the layer attention weights. 
Furthermore, we give a qualitative study to illustrate why our method leads to more accurate results. 
Finally, we evaluate our approach in a closely related task, referring expression comprehension~\cite{DBLP:conf/emnlp/RefCOCO14}, to assess the generality of our approach. 



We summarize the contributions of this paper as follows: 
(1) the first encoder layer fusion approach for the language grounding task;
(2) a novel RSD layer attention mechanism that dynamically aggregates the information in encoder layers for each input visual region;
(3) an extensive empirical analysis of our encoder layer fusion approach; and
(4) new state-of-the-art results in the Talk2Car benchmark. 


\section{Related Work}\label{sec:related-work}

\subsection{Language Grounding for Autonomous Vehicles}\label{sec:related-work-talk2car}
Language grounding for autonomous vehicles~\cite{DBLP:conf/emnlp/talk2car19} is an important task for human-vehicle interactions. To tackle this task, Vandenhende et al.~\cite{DBLP:journals/corr/c4avbaseline} propose the C4AV-Base model that utilizes the bi-directional GRU~\cite{DBLP:conf/emnlp/GRU14} and ResNet models~\cite{DBLP:conf/cvpr/resnet16} to encode the command and the regions respectively and then compute their feature correlation. 
Later, the ASSMR method~\cite{DBLP:conf/eccv/OuZ20} applies an attention mechanism to strengthen the features extracted by bi-directional GRU and ResNet. 
To enhance the reasoning ability, the MSRR method~\cite{DBLP:journals/corr/MSRR2020} introduces a spatial memory module and a multi-step reasoning module to iteratively score the regions. 
Other methods are built on the Transformer~\cite{DBLP:conf/nips/Transformer17}. 
The CMTR method~\cite{DBLP:conf/eccv/LuoDSD20} applies the Transformer encoder-decoder model to model the command and regions separately. 
Dai et al.~\cite{DBLP:conf/eccv/DaiLDS20} use the VL-BERT pre-trained model~\cite{DBLP:conf/iclr/VLBERT20} to jointly learn cross-modal representations for the input. They propose an iterative stacking algorithm, Stack-VL-BERT, to train a deeper VL-BERT model. 
This method only uses the top layer representations of a pre-trained model to compute matching scores, while our approach fuses all encoder layers in a pre-trained model. 

\subsection{Referring Expression Comprehension}
The referring expression comprehension task aims to predict a region in an image according to a direct description of an object~\cite{DBLP:conf/emnlp/RefCOCO14}, e.g., ``\textit{girl on the left}''. In contrast, the command expressions in the language grounding for autonomous vehicles task are more complex and involve both action and object descriptions. Earlier methods~\cite{DBLP:conf/cvpr/RefCOCOg16,DBLP:conf/cvpr/HuXRFSD16,DBLP:conf/cvpr/LuoS17,DBLP:conf/iccv/Liu0017,DBLP:conf/cvpr/Yu0SYLBB18} are built on CNN~\cite{DBLP:journals/neco/CNN89} and LSTM~\cite{DBLP:journals/neco/LSTM97} models to encode the input regions and referring expression. 
Recent methods~\cite{DBLP:conf/eccv/UNITER20,DBLP:conf/emnlp/TanB19,DBLP:conf/iclr/VLBERT20} use V\&L pre-trained models based on the Transformer architecture and they achieve state-of-the-art performance. 
These methods only use the representations in the last layer of a pre-trained model to predict the matching scores, while our method fuses the representations across all encoder layers. 
\subsection{Language Grounding in Human-robot Interaction}
Several methods~\cite{deits2013clarifying, DBLP:journals/jair/ThomasonPSWJYHS20,DBLP:conf/icra/ThomasonPS0JYHS19} allow robots to ask clarification questions and they rely on a semantic parser~\cite{DBLP:journals/jair/ThomasonPSWJYHS20,DBLP:conf/icra/ThomasonPS0JYHS19} or a probabilistic model~\cite{deits2013clarifying} to decompose a command and then ground the resulting constituents to visual regions. 
Shridhar and Hsu~\cite{DBLP:conf/rss/ShridharH18} propose a robot system that picks up an object according to a command. Their method generates a caption for each region and then clusters the generated captions with the input command. 
Kim et al.~\cite{DBLP:conf/cvpr/KimMCTC19} introduce a vehicle controller that accepts advice from humans. Their model uses LSTM, CNN, and visual attention~\cite{DBLP:conf/icml/XuBKCCSZB15} to ground an advice to an image. 
In contrast, our approach is built on V\&L pre-trained models to ground a command to a visual scene. 


\subsection{Encoder Layers Fusion}\label{sec:related-work-layer-fusion}
Encoder layer fusion techniques can be categorized into layer aggregation~\cite{DBLP:conf/cvpr/DeepLayerAgg18,DBLP:conf/aaai/DynamLayerAgg19,DBLP:conf/acl/WangLXZLWC19}, layer-wise coordination~\cite{DBLP:conf/nips/HeTXHQ0L18}, and layer attention~\cite{DBLP:conf/naacl/Elmo18,DBLP:conf/emnlp/BapnaCFCW18}. 
Layer-wise coordination modifies the structure of Transformer and cannot be easily applied to pre-trained models. 
Some layer aggregation methods~\cite{DBLP:conf/aaai/DynamLayerAgg19} dynamically fuse the representations for each input word, but they do not provide normalized weights to interpret the importance of each layer and they introduce millions of new parameters. 
Layer attention utilizes normalized attention weights to fuse encoder layers and allows us to directly interpret the contribution of each layer. 
Existing layer attention methods assign static attention weights to layers independent of the input, whereas our RSD layer attention dynamically predicts attention weights for every input region.

\begin{figure*}[t]
\centering
\includegraphics[width=0.90\linewidth]{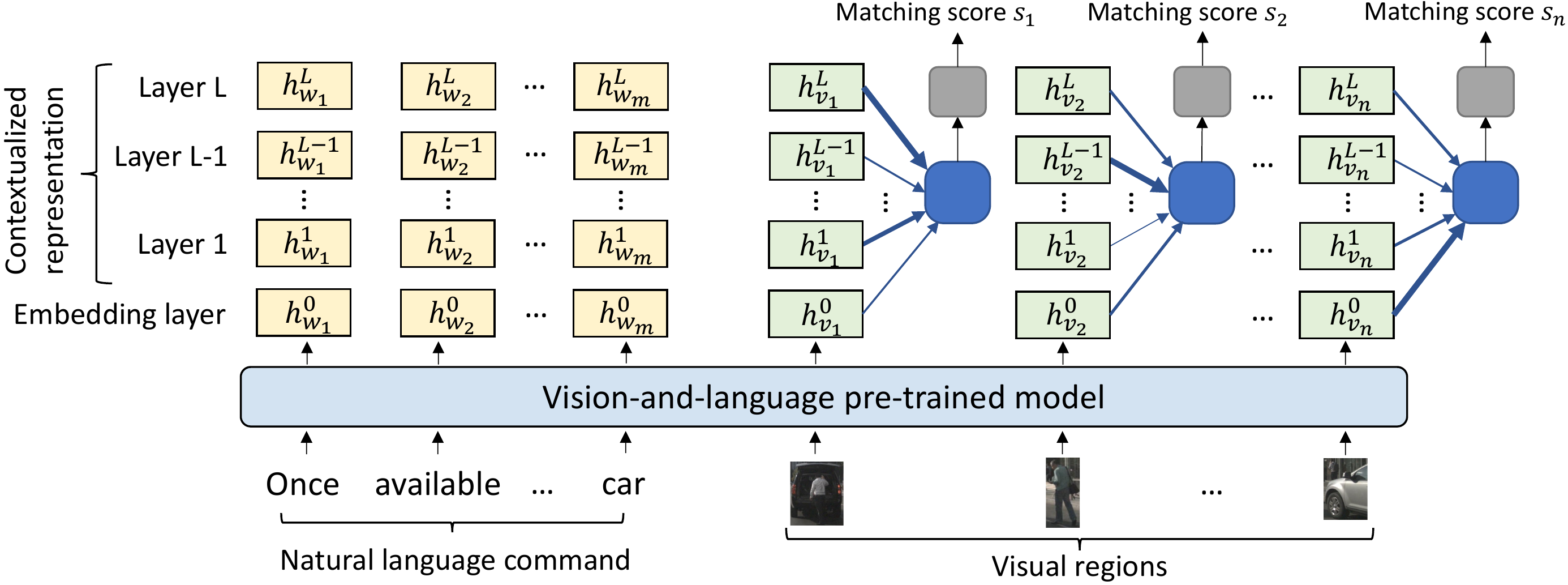}
\caption{Overview of our approach. A V\&L pre-trained model first encodes the command and image regions into multiple layers of representation vectors. Then for each visual region, our RSD layer attention mechanism (indicated by blue rounded square) fuses the representations across encoder layers. Finally, we feed every fused representation to a linear output layer (grey rounded square) to predict the matching scores of regions. 
}
\label{fig:layer-self-attn-fusion}
\end{figure*}

\section{Problem Definition}
Given a natural language command $\mathbf{c}$ and an image $\mathbf{I}$, the goal is to predict the region in the image that the command is referring to. Following previous literature~\cite{DBLP:journals/corr/c4avbaseline,DBLP:conf/iclr/VLBERT20}, we formulate the task as selecting the best-matched region $v^*$ from a set of region proposals $\{v_i\}_{i=1}^{n}$ in the image $\mathbf{I}$ according to the command $\mathbf{c}$. 

\section{Our Layer Fusion Approach}
In our approach, we first pass the input command and image to a V\&L pre-trained model to obtained multiple layers of cross-modal representations. 
We then propose the region-specific dynamic (RSD) layer attention mechanism to dynamically fuse representations across all layers for each individual visual region. The fused representations are then fed to a linear output layer to predict the matching scores of the regions. We display the overall architecture of our approach in Figure~\ref{fig:layer-self-attn-fusion}. 

\subsection{Vision-and-language (V\&L) Pre-trained Model. }
As a preprocessing step, we obtain the region proposals $\{v_i\}_{i=1}^{n}$ of the input image from the Centernet model~\cite{DBLP:journals/corr/centernet19}. We use the WordPieces tokenizer of BERT~\cite{DBLP:conf/naacl/BERT19} to tokenize the input command into a sequence of tokens $w_1, \ldots, w_m$. 
A V\&L pre-trained model takes the region proposals and command tokens as input. 
Inside the pre-trained model, an embedding layer first converts the input tokens and regions into a sequence of word embeddings, $\mathbf{h}_{w_1}^{0},\ldots,\mathbf{h}_{w_m}^{0}$, and region embeddings, $\mathbf{h}_{v_1}^{0},\ldots,\mathbf{h}_{v_n}^{0}$, respectively. 
We refer to the embedding layer as the 0-th encoder layer of the pre-trained model. 


Then, the model feeds the embeddings to a Transformer encoder with $L$ layers. 
At each layer from $1$ to $L$, the encoder uses the multi-head attention mechanism~\cite{DBLP:conf/nips/Transformer17} to model intra-modal and/or cross-modal interactions among the input to produce a $d$-dimensional contextualized representation vector for each region and token. We use $\mathbf{h}_{v_i}^{l}$ to denote the contextualized representation of region $v_i$ produced by the $l$-th encoder layer. Several studies~\cite{DBLP:conf/naacl/Elmo18, DBLP:conf/acl/BertPipeline19} reveal that lower encoder layers extract more concrete features (e.g, position) while higher encoder layers extract more abstract features (e.g., coreference relation). The model is pre-trained on a massive dataset to learn visual and linguistic knowledge. 

There are two major classes of V\&L pre-trained models: (1) single-stream architecture, which models both intra-modal and cross-modal interactions in every encoder layer; and (2) dual-steam architecture, which only allows intra-modal interactions in early encoder layers and then encodes both intra-modal and cross-modal interactions in latter layers. 
We apply the \textbf{UNITER}~\cite{DBLP:conf/eccv/UNITER20} model from single-stream architecture and \textbf{LXMERT}~\cite{DBLP:conf/emnlp/TanB19} from dual-stream architecture because they achieve excellent performance in many V\&L tasks~\cite{DBLP:conf/eccv/CaoGCY0020}. 
On the other hand, SOTA methods~\cite{DBLP:conf/eccv/DaiLDS20} in this task uses another single-stream model, VL-BERT~\cite{DBLP:conf/iclr/VLBERT20}. Their model obtains lower scores than UNITER and LXMERT (see Table~\ref{table:talk2car-main-results} for the results).


\subsection{Region-specific Dynamic (RSD) Layer Attention. }
Prior methods feed the top layer representation of a region $\mathbf{h}_{v_i}^{L}$ to a linear layer to predict its matching score. 
To effectively exploit the concrete and abstract features embedded in different layers, we propose a novel RSD layer attention mechanism to fuse the encoder representations across layers before the prediction of matching scores. 
The overall idea is to dynamically assign attention weights to all encoder layers based on a region's representation vectors in these layers. The weights are then used to aggregate the representation vectors across layers. Our intuition is that every region needs distinct types of features to decide whether it is referred to by the command. 


More concretely, for every region proposal $v_i$, our RSD layer attention method passes its representation $\mathbf{h}_{v_i}^{l}$ at each encoder layer $l$ into a linear layer to compute a relevance score $\alpha^{l}_{i}$. Then we use the softmax function to normalize the relevance scores over all encoder layers and obtain the layer attention weights for region $v_i$, as shown in the following equations: 
\begin{align}\label{eq:layer-self-attn}
    \alpha_{i}^{l} = \mathbf{W}_{\alpha} \mathbf{h}_{v_i}^{l}+b_{\alpha}\text{,} \quad 
    a_{i}^{l} = \frac{\text{exp}(\alpha_{i}^{l})}{\sum_{l'=0}^{L}\text{exp}(\alpha_{i}^{l'})}\text{,}
\end{align}
where $a_{i}^{l}$ denotes the layer attention weight at layer $l$ by region $v_i$ and indicates the importance of the $l$-th layer to region $v_i$. 
In contrast, existing layer attention methods~\cite{DBLP:conf/naacl/Elmo18,DBLP:conf/emnlp/BapnaCFCW18,DBLP:conf/iclr/UnderstandLayerFusion21} only allocate fixed attention weights to encoder layers regardless of the model input. 
We then fuse the representations at different layers by weighted sum: $\tilde{\mathbf{h}}_{v_i}=\sum_{l=0}^{L} a_{i}^{l} \mathbf{h}_{v_i}^{l}$. 
Our RSD layer attention is parameter-efficient since it only introduces $d+1$ new parameters, where $d=768$ for the UNITER and LXMERT models. 
\subsection{Output Layer and Loss Function. }
Finally, we feed the fused encoder representation $\tilde{\mathbf{h}}_{v_i}$ of each region $v_i$ to a linear output layer to predict a matching score: $s_{i}=\sigma(\mathbf{W}_{s}{\tilde{\mathbf{h}}}_{v_i}^{L}+b_{s})$, where $\mathbf{W}_{s}\in \mathbb{R}^{d\times 1}$ and $b_{s}\in \mathbb{R}$, $\sigma$ is the sigmoid function that normalizes the output to the range of $[0,1]$. 
Following previous work~\cite{volta2021}, we use the \textbf{intersection-over-union (IoU)} score between a region $v_i$ and the ground-truth region as the ground-truth matching score $s^*_{i}$. 
The IoU score of a predicted region is the intersection between that region and the ground-truth region divided by the union of them. 
The training objective is the \textbf{binary cross-entropy loss}: $s^*_{i}\log \sigma(s_{i})+(1-s^*_{i})\log(1-\sigma(s_{i}))$. During inference, we use the region proposal that has the highest predicted matching score as the model output. 

\subsection{Model Ablation. }
To verify the importance of region-specific attention weights in our method, we make an ablation in our RSD layer attention to construct a baseline called \textbf{sample-specific layer attention}. Instead of predicting layer attention weights for each individual region, we compute layer attention weights for the entire input sample using the mean-pooled representation of the regions. Then, each region in the sample uses the same set of attention weights to fuse encoder layers. 

Specifically, at each encoder layer $l$, we perform mean-pooling over the representations of all regions: $\mathbf{h}^{l}=1/n \sum_{i=1}^{n}\mathbf{h}_{v_i}^{l}$. Next, we feed the mean-pooled representation $\mathbf{h}^{l}$ to a linear layer to learn a relevance score. All the relevance scores are then normalized by softmax to yield the layer attention weights shared by all the regions.

\section{Experimental Setup}
\subsection{Datasets}
We use the \textbf{Talk2Car}~\cite{DBLP:conf/emnlp/talk2car19} dataset to conduct our experiments. Talk2Car is the standard benchmark for the language grounding for autonomous vehicles task. The images are urban scenes captured by cameras on a car. Each input text is a natural language command referring to a particular object in the input image, e.g., ``\textit{get a parking spot near the second car on the left side}''. A command expression contains both an action instruction and an object description. The averaged length of expressions is 11.0. The training, validation, and test splits contain 8,349/1,163/2,447 samples. 

We further evaluate our method on the \textbf{RefCOCO+}~\cite{DBLP:conf/emnlp/RefCOCO14} and \textbf{RefCOCOg} \cite{DBLP:conf/cvpr/RefCOCOg16} datasets, which are popular benchmarks for referring expression comprehension. The input text in these two datasets is a description of an object in the image (e.g., ``\textit{giraffe with lowered head}'') rather than a command. 
The average expression length of RefCOCO+ and RefCOCOg datasets are 3.5 and 8.4 respectively. The images in these datasets are collected from the MSCOCO benchmark~\cite{DBLP:conf/eccv/MSCOCO14}. For RefCOCO+, we divide the data according to Bugliarello et al.~\cite{volta2021} and obtain the split of 287,113/13,368/11,490 for training, validation, and test. For RefCOCOg, we use the UMD split of 42,226/2,573/5,023. 






\subsection{Evaluation Metric}
During evaluation, we compute the IoU score of the predicted region. If the IoU score is larger than 0.5, we consider the prediction correct. Following \cite{DBLP:conf/emnlp/talk2car19}, we report the averaged number of correct predictions and refer to it as the IoU$_{0.5}$ score. 

\subsection{Baselines and Comparison}
We adopt recent methods of language grounding for autonomous vehicles as baselines, including C4AV-Base~\cite{DBLP:journals/corr/c4avbaseline}, 
ASSMR~\cite{DBLP:conf/eccv/OuZ20}, MSRR~\cite{DBLP:journals/corr/MSRR2020}, CMTR~\cite{DBLP:conf/eccv/LuoDSD20}, and Stack-VL-BERT~\cite{DBLP:conf/eccv/DaiLDS20}. 
Moreover, we consider the UNITER~\cite{DBLP:conf/eccv/UNITER20} and LXMERT~\cite{DBLP:conf/emnlp/TanB19} models as baselines. 
For the UNITER model, we adopt the base model with 12 layers. 
Furthermore, we compare with previous layer fusion methods: 
\textbf{coarse-grained layer attention}~\cite{DBLP:conf/naacl/Elmo18,DBLP:conf/emnlp/BapnaCFCW18}, which assigns static attention weights to encoder layers, 
\textbf{fine-grained layer attention}~\cite{DBLP:conf/iclr/UnderstandLayerFusion21}, which assigns static attention weights to the elements of the representations in all layers, 
\textbf{dynamic combination}~\cite{DBLP:conf/aaai/DynamLayerAgg19}, which uses $L$ feedforward networks to aggregate the representations in different layers, 
\textbf{dynamic routing}~\cite{DBLP:conf/aaai/DynamLayerAgg19}, which iteratively refines the fused representation based on the agreement between each layer representation and the fused representation. 
We use \textbf{RSD-UNITER} and \textbf{RSD-LXMERT} to denote the UNITER and LXMERT models after being enhanced by our layer fusion approach.

\begin{table}[t]
\centering
\caption{
IoU$_{0.5}$ scores of different models on the Talk2Car dataset. We bold the best results and underline the second-best results. 
}
\label{table:talk2car-main-results}
\begin{tabular}{l|cc}
\hline \hline
\textbf{Model}              &  \textbf{Val. set}  & \textbf{Test set}   \\
\hline \hline
C4AV-Base          & 43.5 & 44.1 \\ 
MAC          & - & 50.5 \\ 
MSRR          & 60.3 & 60.1 \\ 
ASSMR         & 67.9 & 66.4 \\ 
CMTR          & 68.2 & 69.1 \\ 
Stack-VL-BERT  & 68.2 & 71.0 \\ 
LXMERT        & 72.7 & 73.1 \\ 
RSD-LXMERT (Ours)  & \underline{74.7} & \underline{73.7} \\ 
UNITER        & 74.0 & 73.2 \\ 
RSD-UNITER (Ours)  & \textbf{74.9} & \textbf{73.9} \\ \hline
\end{tabular}
\end{table}

\begin{figure*}[t]
\centering
\begin{subfigure}[t]{\columnwidth}
    \centering
    \includegraphics[width=\columnwidth]{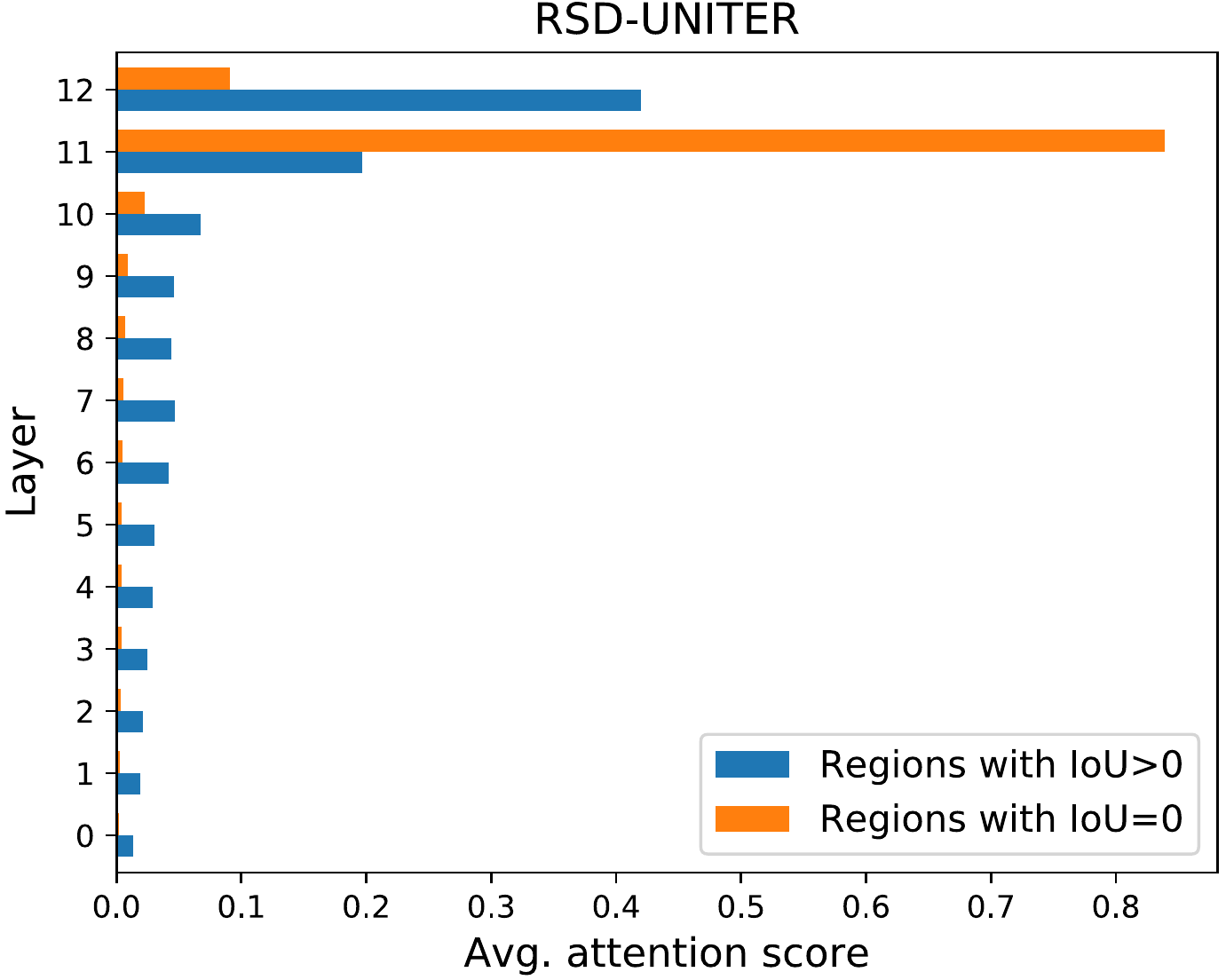}
\end{subfigure}
\begin{subfigure}[t]{\columnwidth}
    \centering
    \includegraphics[width=\columnwidth]{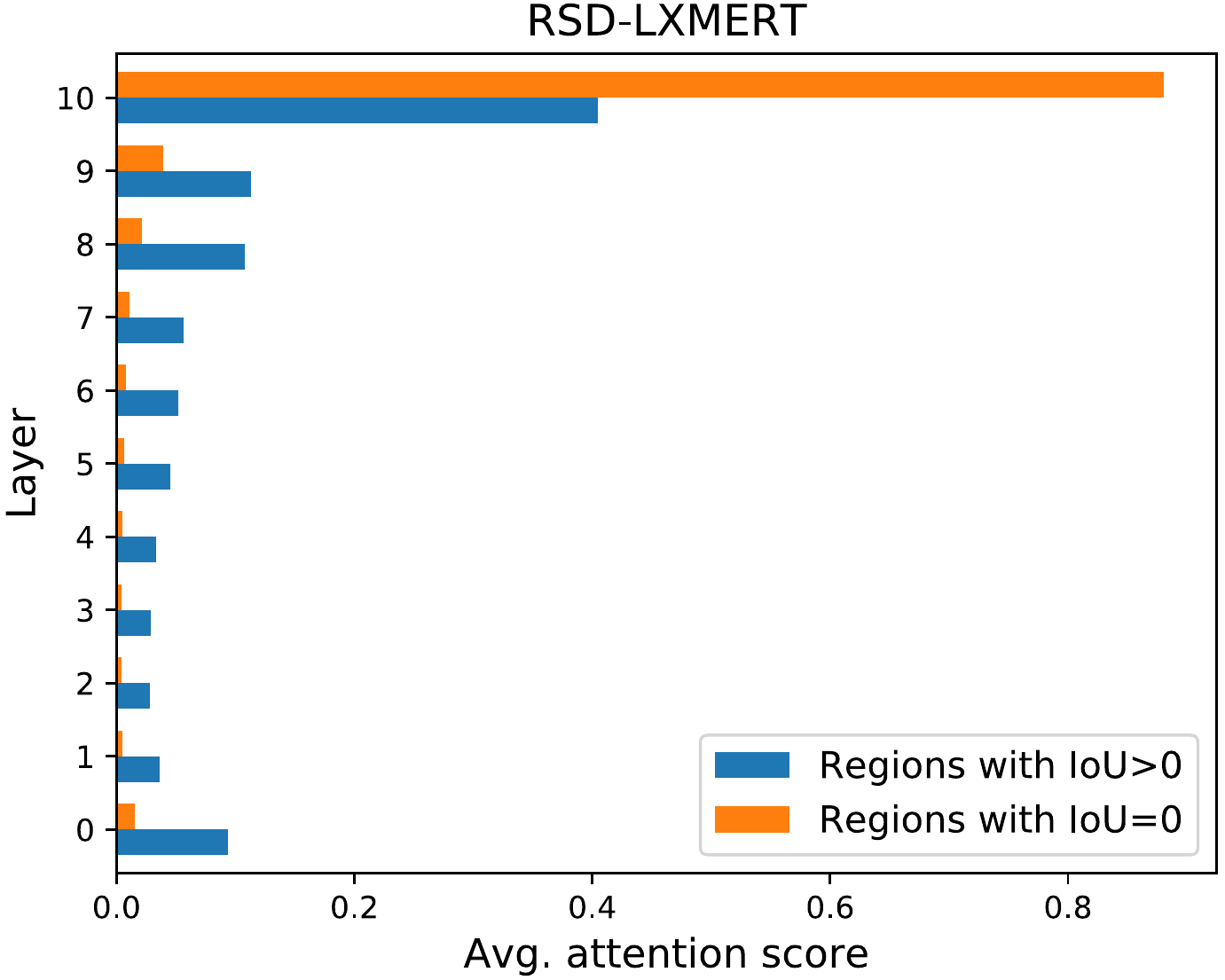}
\end{subfigure}
\caption{Attention weights over all encoder layers learned by our RSD layer attention. The attention weights are averaged over all regions in the validation set of Talk2Car. 
The 0-th layer indicates the embedding layer. 
}
\label{fig:layer-attn-distribution}
\end{figure*}

\section{Experiment Results}
\subsection{Comparison Results with SOTA methods}

We report the main comparison results on the Talk2Car dataset in Table~\ref{table:talk2car-main-results}. 
It is observed that our layer fusion approach significantly and consistently improves the IOU$_{0.5}$ scores of both LXMERT and UNITER models in validation and test sets, which show that our RSD layer attention can increase the accuracy of both single-stream and dual-stream V\&L pre-trained models. 
The reason is that our approach is agnostic to the architecture of the underlying pre-trained model. 
Moreover, both of our methods (RSD-LXMERT and RSD-UNITER) outperform the state-of-the-art models in this benchmark. The above results indicate that it is important to effectively utilize the features embedded in different layers of a V\&L pre-trained model to locate the referred region.

\subsection{Comparison of Different Layer Fusion Methods}
We compare the performance of different encoder layer fusion methods on UNITER since it is the most accurate pre-trained model as shown in the previous section. 
From Table~\ref{table:layer-attention-analysis}, we observe that our RSD layer attention mechanism achieves better performance than existing layer fusion methods. Moreover, we can see that our constructed baseline, sample-specific layer attention, obtains lower IoU$_{0.5}$ scores than our RSD layer attention, which indicate that it is crucial to learn a distinct set of layer attention weights for each visual region. 
Surprisingly, our RSD layer attention substantially outperforms previous dynamic layer fusion methods (rows 4\&5) that introduce millions of new parameters. 
The results suggest that the combination of region specific layer attention weights and weighted sum fusion operation is effective in this task. Previous dynamic aggregation methods may introduce unnecessary parameters which make the model more difficult to train.

\begin{table}[t]
\centering
\caption{
IoU$_{0.5}$ scores of different layer fusion methods in the UNITER model on the Talk2Car dataset. \# Param. denotes the number of parameters. 
}
\label{table:layer-attention-analysis}
\begin{tabular}{c|l|ccr}
\hline \hline
\textbf{\#} & \textbf{Model}              &  \textbf{Val}  & \textbf{Test} & \textbf{\# Param.}  \\
\hline \hline
1 & UNITER        & 74.0 & 73.2 & 112.0M \\ \hline
2 & FineGrainedAttn-UNITER       & 73.5 & 73.0 & +9984 \\ 
3 & CoarseGrainedAttn-UNITER     & 74.0 & 73.6 & +13 \\ \hline
4 & DynamicCombin-UNITER     & 74.1 & 72.3 & +107.3M \\
5 & DynamicRouting-UNITER     & 72.4 & 72.2 & +7.7M \\ \hline
6 & SampleSpecificAttn-UNITER     & 74.6 & 72.6 & +769\\ 
7 & RSD-UNITER   & \textbf{74.9} & \textbf{73.9} & +769 \\ \hline
\end{tabular}
\vspace{-0.1in}
\end{table}

\begin{figure*}[t]
\centering
\begin{subfigure}[t]{\columnwidth}
    \centering
    \includegraphics[width=\columnwidth]{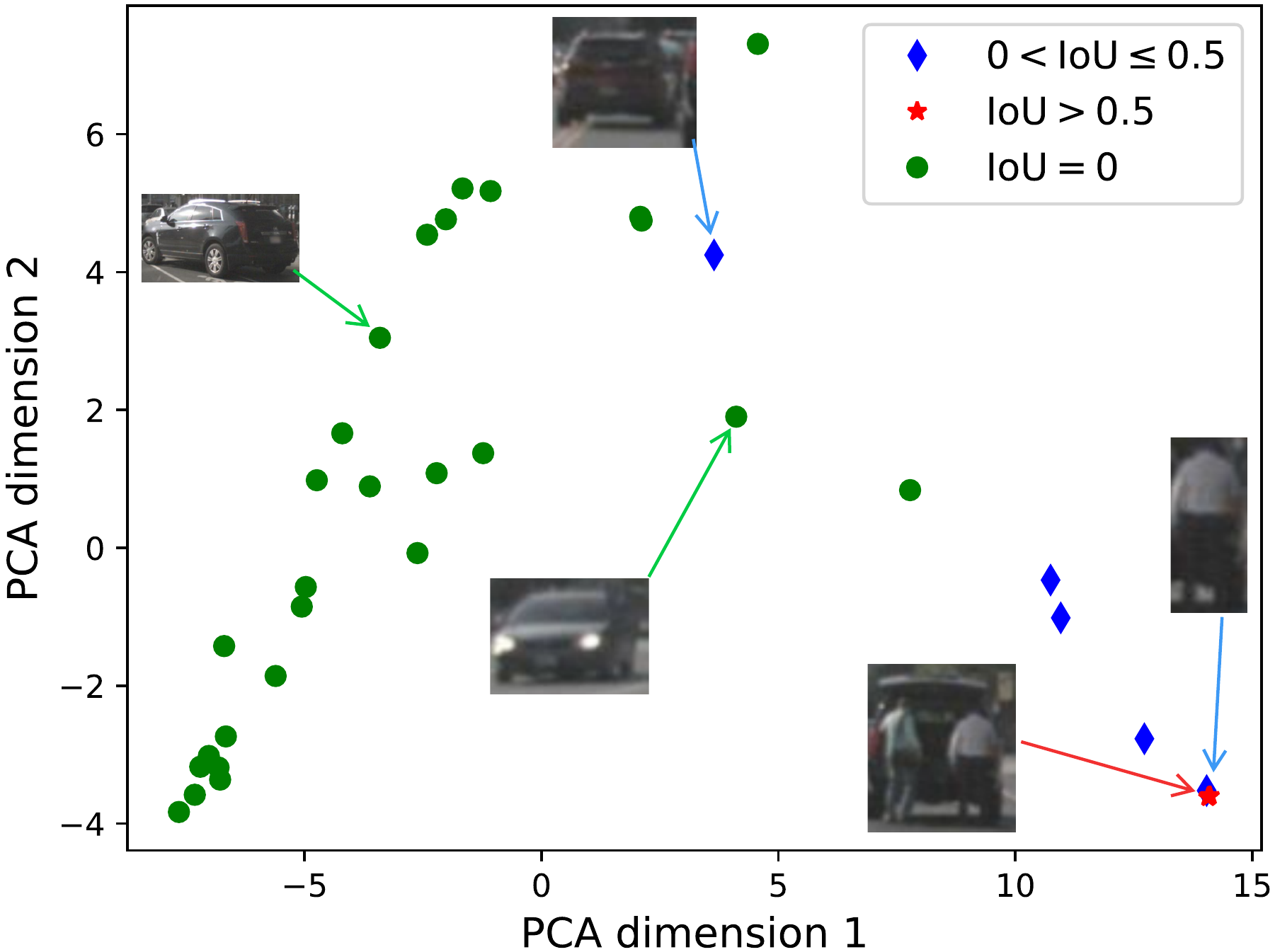}
    \caption{Representations in the 12-th layer of RSD-UNITER. }
    \label{fig:uniter-layer-transformation-12}
\end{subfigure}
\begin{subfigure}[t]{\columnwidth}
    \centering
    \includegraphics[width=\columnwidth]{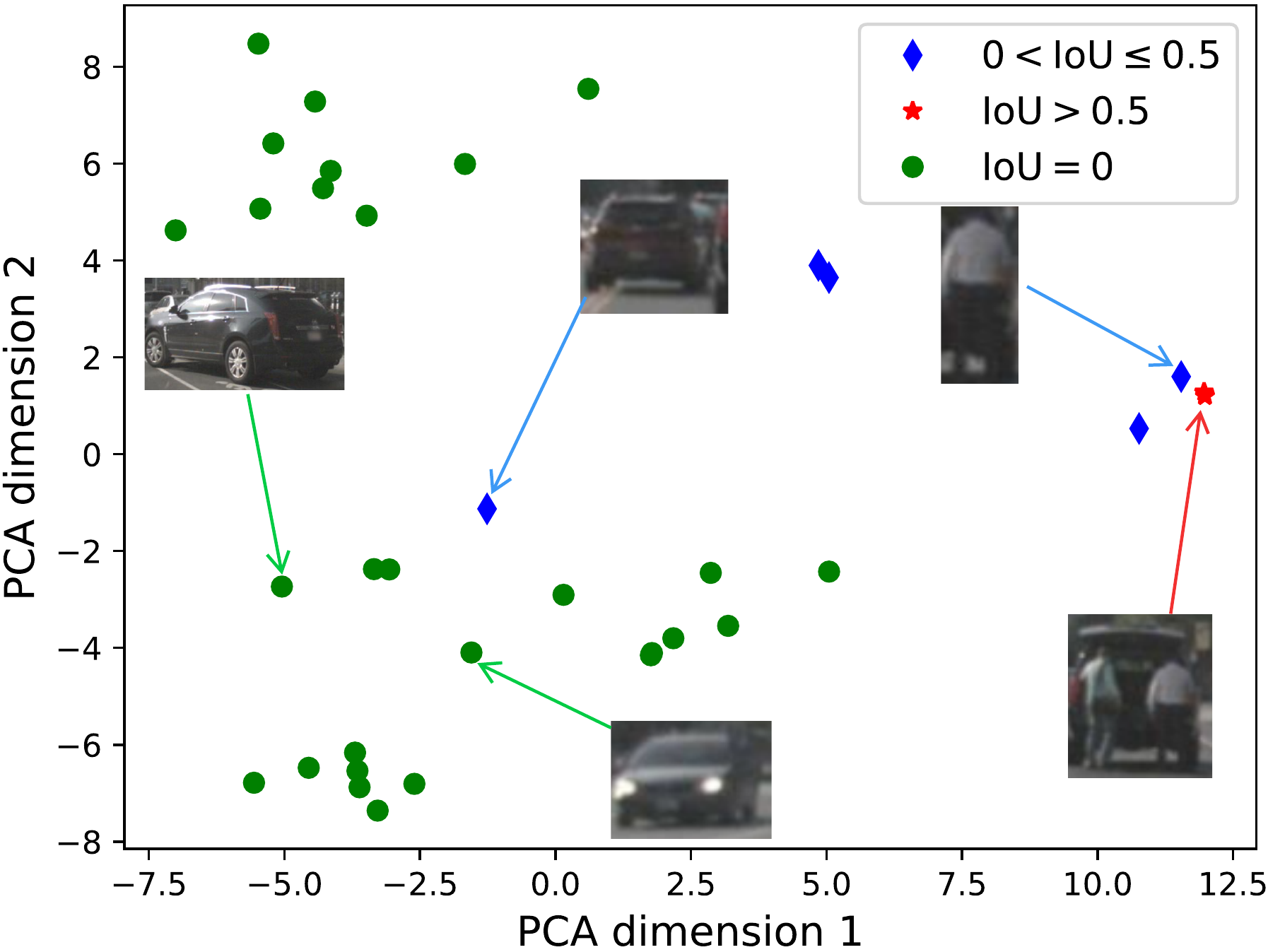}
    \caption{Representations after layer fusion of RSD-UNITER. }
    \label{fig:uniter-layer-transformation-fused}
\end{subfigure}
\caption{Regions' representations of a validation sample in the RSD-UNITER model after projected by PCA. The input command is ``\textit{when it is safe, slowly pass the car that is standing still up ahead}''. Red star denotes the ground-truth region. 
}
\label{fig:uniter-layer-transformation}
\end{figure*}

\subsection{Analysis of Layer Attention Weights}\label{sec:layer-attn-score-analysis}
We analyze the distribution of layer attention weights learned by our RSD layer attention for different regions. 
We partition the input regions into the following two groups according to their IoU with the ground-truth region. {$\text{IoU}>0$}: contains all the regions that have IoU greater than 0 (i.e., the regions that overlap the ground-truth region). {$\text{IoU}=0$}: includes all the regions with IoU equals to 0. 
Figure~\ref{fig:layer-attn-distribution} shows the distribution of layer attention weights in two groups of regions. We draw the following two observations. 
\begin{itemize}
\item \textit{Both models allocate a larger proportion of attention weights to higher encoder layers.} It is because the representations in higher encoder layers are exposed to more cross-modal interactions. Thus, they usually capture more cross-modal semantic features which are essential to command and scene understanding. 
\item \textit{Compared with the regions with $\text{IoU}=0$, the regions with $\text{IoU}>0$ assign more attention weights to lower encoder layers.} 
It is because most of the regions with $\text{IoU}=0$ are obviously irrelevant to the command. Hence, the abstract features embedded in higher encoder layers can provide sufficient information for the model to predict a matching score of zero. On the other hand, the regions with $\text{IoU}>0$ overlap the target region and they are more difficult to be disambiguated. Hence, the model requires more surface-level features (e.g., position) of the regions in lower encoder layers to decide their matching scores. 
\end{itemize}

\subsection{Qualitative Analysis of Region Representations}

To illustrate how our layer fusion method results in more accurate predictions, we qualitatively analyze the representations of visual regions. 
We first collect the regions' representations learned by our RSD-UNITER model. 
We then utilize the Principal Component Analysis (PCA)~\cite{journal/pca1901} technique to project the representation vectors into a 2-dimensional vector space for the sake of visualization. 


Figure~\ref{fig:uniter-layer-transformation} visualizes the regions' representations learned by our RSD-UNITER model.
The ground-truth region's representation is denoted by a red star. 
From Figure~\ref{fig:uniter-layer-transformation-12}, we observe that in the top layer, the ground-truth region's representation is far away from the representations of regions that do not overlap the ground-truth ($\text{IoU}=0$). 
Previous methods feed the top layer representation of a region to a linear layer to predict its matching score. 
However, the ground-truth region is extremely close to a region that has intersection with the ground-truth ($0<\text{IoU}\leq 0.5$). It is difficult for a linear layer to separate the target region from its closest neighbor. 
Then in Figure~\ref{fig:uniter-layer-transformation-fused}, we observe that after applying our RSD layer attention to fuse the representations across layers, the ground-truth region's representation is pushed slightly further from its closest neighbor. Thus, a linear layer can separate out the ground-truth region more easily, which demonstrates the advantage of our approach. 

\subsection{Results in Referring Expression Comprehension}
We further evaluate the performance of our approach on the referring expression comprehension task. 
Table~\ref{table:refcoco-results} presents the results in the RefCOCO+ and RefCOCOg datasets. 
We observe that our approach increases the performance of UNITER and LXMERT models in most of the cases but the improvements are less significant than that in the Talk2Car dataset. 
We analyze the reason as follow. 
The command expressions in the autonomous driving setting are complicated and contain both action and object descriptions. 
The model requires features from multiple layers of cross-modal representations to learn the alignment between text and image. Thus, our encoder layer fusion approach significantly improves the accuracy of predicted regions. 
On the other hand, the referring expressions in RefCOCO+ and RefCOCOg datasets are straight-forward descriptions of a target object. Hence, the top layer representations often capture enough information to find the target object and encoder layer fusion provides a smaller benefit to the model. 


\begin{table}[t]
\centering
\caption{
IoU$_{0.5}$ results in the RefCOCO+ and RefCOCOg datasets. 
}
\label{table:refcoco-results}
\begin{tabular}{l|cccc}
\hline \hline
      & \multicolumn{2}{c}{\textbf{RefCOCO+}} & \multicolumn{2}{c}{\textbf{RefCOCOg}} \\
\textbf{Model}  &  \textbf{Val}  & \textbf{Test}  &  \textbf{Val}  & \textbf{Test}  \\
\hline \hline
UNITER        & 72.0 & 70.8 & 73.1 & 73.2 \\ 
RSD-UNITER   & 72.6 & 71.5 & 74.0 & 73.0 \\ \hline
LXMERT        & 71.3 & 70.0 & 71.9 & 72.1\\ 
RSD-LXMERT   & 71.4 & 70.1 & 72.3 & 72.4\\ \hline
\end{tabular}
\end{table}

\section{Conclusion}
In this work, we present the first encoder layer fusion approach for the language grounding for autonomous vehicles task. 
In our approach, we apply a V\&L pre-trained model to learn contextualized representations for the input command and regions. Then we propose a novel RSD layer attention mechanism to dynamically aggregate the representations across all encoder layers for each individual region proposal. 
Experiment results on the real-world Talk2Car benchmark show that our approach consistently improves the performance of UNITER and LXMERT pre-trained models and achieves the new SOTA results in this task. 






\section*{ACKNOWLEDGMENT}
This paper is supported by National Key Research and Development Program of China (No. 2019YFB2102100), the Science and Technology Development Fund of Macau SAR (File no. 0015/2019/AKP), and Guangdong-Hong Kong-Macao Joint Laboratory of Human-Machine Intelligence-Synergy Systems (No. 2019B121205007). 



\bibliography{IEEEexample.bib}
\bibliographystyle{IEEEtran}

\end{document}